\documentclass[11pt,letterpaper]{article}
\usepackage{ijcnlp2017}
\usepackage{times}
\usepackage{latexsym}
\usepackage{amsfonts}
\usepackage{multirow}
\usepackage{graphicx}
\usepackage{amssymb}
\usepackage{amsmath}
\usepackage{url}

\ijcnlpfinalcopy

\newcommand\BibTeX{B{\sc ib}\TeX}

\title{Speaker Role Contextual Modeling for Language Understanding\\ and Dialogue Policy Learning}

\author{Ta-Chung Chi$^\star$\quad Po-Chun Chen$^\star$\quad Shang-Yu Su$^\dagger$\quad Yun-Nung Chen$^\star$\\
$^\star$Department of Computer Science and Information Engineering\\
$^\dagger$Graduate Institute of Electrical Engineering\\
National Taiwan University\\
\texttt{ \{b02902019,r06922028,r05921117\}@ntu.edu.tw\quad y.v.chen@ieee.org}
}

\date{}

\begin{document}

\maketitle

\begin{abstract}
Language understanding (LU) and dialogue policy learning are two essential components in conversational systems.
Human-human dialogues are not well-controlled and often random and unpredictable due to their own goals and speaking habits.
This paper proposes a role-based contextual model to consider different speaker roles independently based on the various speaking patterns in the multi-turn dialogues.
The experiments on the benchmark dataset show that the proposed role-based model successfully learns role-specific behavioral patterns for contextual encoding and then significantly improves language understanding and dialogue policy learning tasks\footnote{The source code is available at: \url{https://github.com/MiuLab/Spk-Dialogue}.}.
\end{abstract}

\section{Introduction}
\label{sec:intro}
Spoken dialogue systems that can help users to solve complex tasks such as booking a movie ticket become an emerging research topic in the artificial intelligence and natural language processing area. 
With a well-designed dialogue system as an intelligent personal assistant, people can accomplish certain tasks more easily via natural language interactions. 
Today, there are several virtual intelligent assistants, such as Apple's Siri, Google's Home, Microsoft's Cortana, and Amazon's Echo. Recent advance of deep learning has inspired many applications of neural models to dialogue systems. \newcite{wen2016network}, \newcite{bordes2017learning}, and \newcite{li2017end} introduced network-based end-to-end trainable task-oriented dialogue systems.

A key component of the understanding system is a language understanding (LU) module---it parses user utterances into semantic frames that capture the core meaning, where three main tasks of LU are domain classification, intent determination, and slot filling~\cite{tur2011spoken}.
A typical pipeline of LU is to first decide the domain given the input utterance, and based on the domain, to predict the intent and to fill associated slots corresponding to a domain-specific semantic template.
Recent advance of deep learning has inspired many applications of neural models to natural language processing tasks. With the power of deep learning, there are emerging better approaches of LU~\cite{hakkani2016multi,chen2016knowledge,chen2016syntax,wang2016learning}.
However, most of above work focused on single-turn interactions, where each utterance is treated independently.

\begin{figure*}[t]
\centering
\includegraphics[width=\linewidth]{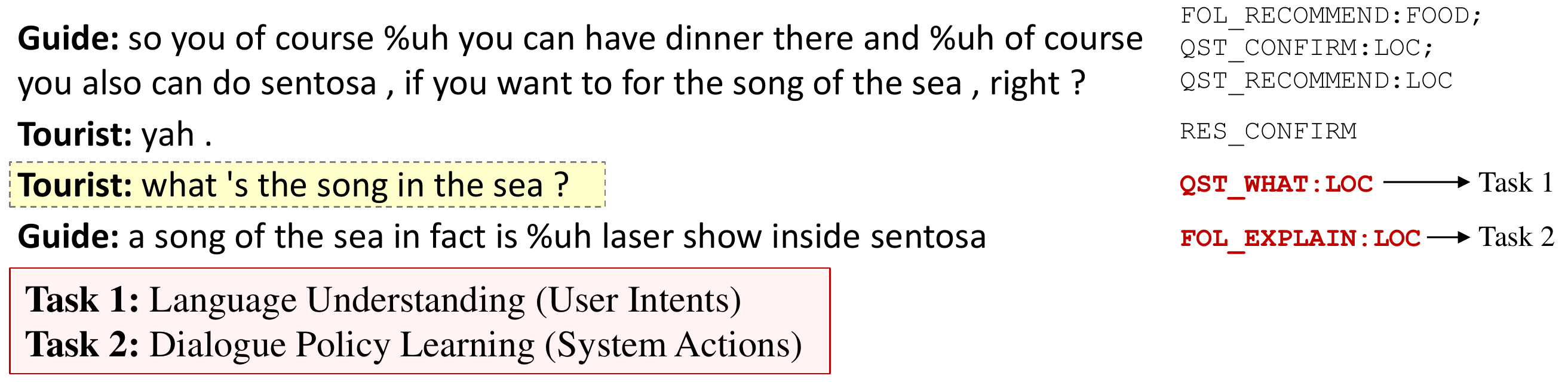}
\vspace{-3mm}
\caption{The human-human conversational utterances and their associated semantics from DSTC4.}
\label{fig:example}
\end{figure*}

The contextual information has been shown useful for LU~\cite{bhargava2013easy,xu2014contextual,chen2015leveraging,sun2016an}.
For example, the Figure~\ref{fig:example} shows conversational utterances, where the intent of the highlighted tourist utterance is to ask about location information, but it is difficult to understand without contexts.
Hence, it is more likely to estimate the location-related intent given the contextual utterance about location recommendation.
Contextual information has been incorporated into the recurrent neural network (RNN) for improved domain classification, intent prediction, and slot filling~\cite{xu2014contextual,shi2015contextual,weston2015memory,chen2016end}.
The LU output is semantic representations of users' behaviors, and then flows to the downstream dialogue management component in order to decide which action the system should take next, as called \emph{dialogue policy}.
It is intuitive that better understanding could improve the dialogue policy learning, so that the dialogue management can be further boosted through interactions~\cite{li2017end}.

Most of previous dialogue systems did not take speaker roles into consideration.
However, we discover that different speaker roles can cause notable variance in speaking habits and later affect the system performance differently~\cite{chen2017dynamic}. 
From Figure~\ref{fig:example}, the benchmark dialogue dataset, Dialogue State Tracking Challenge 4 (DSTC4)~\cite{kim2016fourth}\footnote{\url{http://www.colips.org/workshop/dstc4/}}, contains two specific roles, a tourist and a guide.
Under the scenario of dialogue systems and the communication patterns, we take the tourist as a user and the guide as the dialogue agent (system).
During conversations, the user may focus on not only \emph{reasoning (user history)} but also \emph{listening (agent history)}, so different speaker roles could provide various cues for better understanding and policy learning.

This paper focuses on LU and dialogue policy learning, which targets the understanding of tourist's natural language (LU; language understanding) and the prediction of how the system should respond (SAP; system action prediction) respectively.
In order to comprehend what the tourist is talking about and predict how the guide reacts to the user, this work proposes a role-based contextual model by modeling role-specific contexts differently for improving system performance.

\begin{figure*}[t]
\centering
\includegraphics[width=\linewidth]{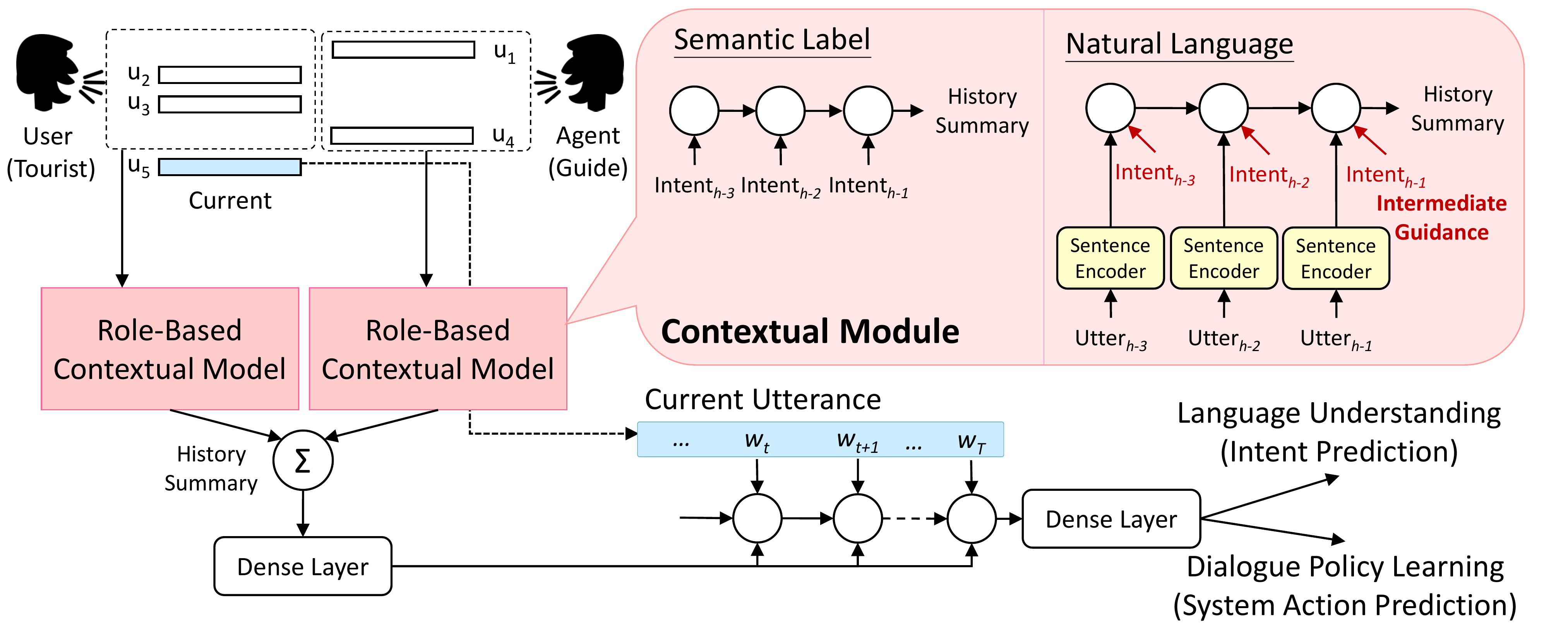}
\vspace{-3mm}
\caption{Illustration of the proposed role-based contextual model.}
\label{fig:model}
\end{figure*}

\section{Proposed Approach}
\label{sec:model}

The model architecture is illustrated in Figure~\ref{fig:model}.
First, the previous utterances are fed into the contextual model to encode into the history summary, and then the summary vector and the current utterance are integrated for helping LU and dialogue policy learning.
The whole model is trained in an end-to-end fashion, where the history summary vector is automatically learned based on two downstream tasks.
The objective of the proposed model is to optimize the conditional probability $p(\mathbf{\hat{y}}\mid \mathbf{x})$, so that the difference between the predicted distribution $q(\hat{y_k}=z\mid \mathbf{x})$ and the target distribution $q(y_k=z\mid \mathbf{x})$ can be minimized:
\begin{equation}
\mathcal{L}=-\sum_{k=1}^{K}\sum_{z=1}^{N}q(y_k=z\mid \mathbf{x}) \log p(\hat{y_k}=z\mid \mathbf{x}),
\end{equation}
where the labels $\textbf{y}$ can be either intent tags for understanding or system actions for dialogue policy learning.

\paragraph{Language Understanding (LU)}
Given the current utterance $\textbf{x}=\{w_t\}^T_1$, the goal is to predict the user intents of $\textbf{x}$, which includes the speech acts and associated attributes shown in Figure~\ref{fig:example}; for example, \texttt{QST\_WHAT} is composed of the speech act \texttt{QST} and the associated attribute \texttt{WHAT}.
Note that we do not process the slot filling task for extracting \texttt{LOC}.
We apply a bidirectional long short-term memory (BLSTM) model~\cite{schuster1997bidirectional} to integrate preceding and following words to learn the probability distribution of the user intents.
\begin{eqnarray}
\label{eq:basic}
\textbf{v}_\text{cur} &=& \text{BLSTM}(\textbf{x}, W_\text{his}\cdot \textbf{v}_\text{his}),\\
\textbf{o} &=& \mathtt{sigmoid}(W_\text{LU}\cdot \textbf{v}_\text{cur}),
\end{eqnarray}
where $W_\text{his}$ is a dense matrix and $\textbf{v}_\text{his}$ is the history summary vector, $\textbf{v}_\text{cur}$ is the context-aware vector of the current utterance encoded by the BLSTM, and $\textbf{o}$ is the intent distribution.
Note that this is a multi-label and multi-class classification, so the $\mathtt{sigmoid}$ function is employed for modeling the distribution after a dense layer.
The user intent labels $\textbf{y}$ are decided based on whether the value is higher than a threshold $\theta$.

\paragraph{Dialogue Policy Learning}
For system action prediction, we also perform similar multi-label multi-class classification on the context-aware vector $\textbf{v}_\text{cur}$ from (\ref{eq:basic}) using $\mathtt{sigmoid}$:
\begin{equation}
\textbf{o} = \mathtt{sigmoid}(W_\pi \cdot \textbf{v}_\text{cur}),
\end{equation}
and then the system actions can be decided based on a threshold $\theta$.

\subsection{Contextual Module}
\label{ssec:contexualmodel}

In order to leverage the contextual information, we utilize two types of contexts: 1) semantic labels and 2) natural language, to learn history summary representations, $\textbf{v}_\text{his}$ in (\ref{eq:basic}).
The illustration is shown in the top-right part of Figure~\ref{fig:model}.

\paragraph{Semantic Label}
Given a sequence of annotated intent tags and associated attributes for each history utterance, we employ a BLSTM to model the explicit semantics:
\begin{equation}
\textbf{v}_\text{his} = \text{BLSTM}(\text{intent}_t),
\label{eq:tag}
\end{equation}
where $\text{intent}_t$ is the vector after one-hot encoding for representing the annotated intent and the attribute features.
Note that this model requires the ground truth annotations of history utterances for training and testing.

\begin{figure}[t]
\centering
\includegraphics[width=\linewidth]{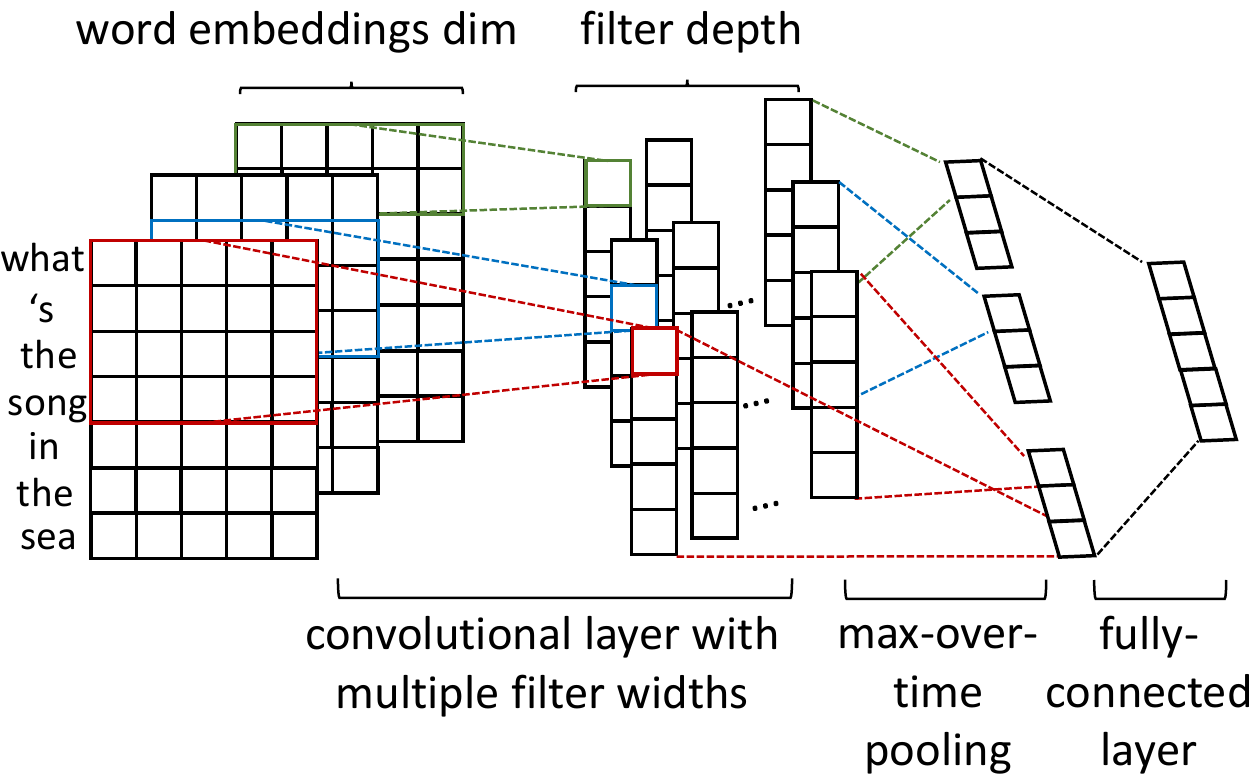}
\vspace{-3mm}
\caption{Illustration of the CNN sentence encoder for the example sentence ``\textit{what's the song in the sea}''.}
\label{fig:cnn}
\end{figure}

\paragraph{Natural Language (NL)}
Given the natural language history, a sentence encoder is applied to learn a vector representation for each prior utterance.
After encoding, the feature vectors are fed into a BLSTM to capture temporal information:
\begin{equation}
\textbf{v}_\text{his} = \text{BLSTM}(\text{CNN}(\text{utt}_t)),
\label{eq:nl}
\end{equation}
where the CNN is good at extracting the most salient features that can represent the given natural language utterances illustrated in Figure~\ref{fig:cnn}.
Here the sentence encoder can be replaced into different encoders\footnote{In the experiments, CNN achieved slightly better performance with fewer parameters compared with BLSTM.}, and the weights of all encoders are tied together.

\paragraph{NL with Intermediate Guidance}
Considering that the semantic labels may provide rich cues, the middle supervision signal is utilized as intermediate guidance for the sentence encoding module in order to guide them to project from input utterances to a more meaningful feature space.
Specifically, for each utterance, we compute the cross entropy loss between the encoder outputs and corresponding intent-attributes shown in Figure~\ref{fig:model}.
Assuming that $l_t$ is the encoding loss for $\text{utt}_t$ in the history, the final objective is to minimize $(\mathcal{L}+\sum_t{l_t})$.
This model does not require the ground truth semantics for history when testing, so that it is more practical compared to the above model using semantic labels.

\subsection{Speaker Role Modeling}
\label{ssec:rolebasedmodel}

In a dialogue, there are at least two roles communicating with each other, each individual has his/her own goal and speaking habit.
For example, the tourists have their own desired touring goals and the guides are try to provide the sufficient touring information for suggestions and assistance.
Prior work usually ignored the speaker role information or only modeled a single speaker's history for various tasks~\cite{chen2016end,yang2017end}.
The performance may be degraded due to the possibly unstable and noisy input feature space.
To address this issue, this work proposes the role-based contextual model: instead of using only a single BLSTM model for the history, we construct one individual contextual module for each speaker role.
Each role-dependent recurrent unit $\text{BLSTM}_{\text{role}_x}$ receives corresponding inputs $\text{x}_{i,\text{role}_x}$ ($i=[1, ..., N]$), which have been processed by an encoder model, we can rewrite (\ref{eq:tag}) and (\ref{eq:nl}) into (\ref{eq:tag2}) and (\ref{eq:nl2}) respectively:
\begin{eqnarray}
\label{eq:tag2}
\textbf{v}_\text{his} &=& \text{BLSTM}_{\text{role}_a}(\text{intent}_{t,\text{role}_a}) \\
&+& \text{BLSTM}_{\text{role}_b}(\text{intent}_{t,\text{role}_b}).\nonumber\\
\label{eq:nl2}
\textbf{v}_\text{his} &=& \text{BLSTM}_{\text{role}_a}(\text{CNN}(\text{utt}_{t,\text{role}_a}))\\\nonumber
&+& \text{BLSTM}_{\text{role}_b}(\text{CNN}(\text{utt}_{t,\text{role}_b}))
\end{eqnarray}
Therefore, each role-based contextual module focuses on modeling the role-dependent goal and speaking style, and $\textbf{v}_\text{cur}$ from (\ref{eq:basic}) is able to carry role-based contextual information.

\begin{table*}
\centering
\begin{tabular}{ |l  l l | c | c | c | }
    \hline
    \multicolumn{3}{|c|}{\multirow{2}{*}{\bf Model}} & \bf Language & \bf Policy \\
    &  & & \bf Understanding & \bf Learning\\\hline\hline
   Baseline & (a) & \emph{DSTC4-Best} & 52.1~ & -\\
   & (b) & BLSTM  & 62.6~ & 63.4~~\\\hline
   Contextual-Sem & (c) & BLSTM & 68.2~ & 66.8~~\\
   & (d) & ~+ Role-Based  & \bf 69.2$^\dag$ & \bf 70.1$^\dag$\\\hline
   Contextual-NL & (e) & BLSTM & 64.2~ & 66.3~~\\
   & (f) & ~+ Role-Based   & 65.1$^\dag$ & 66.9$^\dag$\\
   & (g) & ~+ Role-Based w/ Intermediate Guidance &  \bf  65.8$^\dag$ & \bf 67.4$^\dag$\\
    \hline
  \end{tabular}
  \vspace{-1mm}
\caption{Language understanding and dialogue policy learning performance of F-measure on DSTC4 (\%). $^\dag$ indicates the significant improvement compared to all methods without speaker role modeling.}
\label{tab:res}
\end{table*}

\section{Experiments}
\label{sec:experiments}

To evaluate the effectiveness of the proposed model, we conduct the LU and dialogue policy learning experiments on human-human conversational data. 


\subsection{Setup}
\label{ssec:settings}
The experiments are conducted on DSTC4, which consists of 35 dialogue sessions on touristic information for Singapore collected from Skype calls between 3 tour guides and 35 tourists~\cite{kim2016fourth}. 
All recorded dialogues with the total length of 21 hours have been manually transcribed and annotated with speech acts and semantic labels at each turn level.
The speaker labels are also annotated.
Human-human dialogues contain rich and complex human behaviors and bring much difficulty to all dialogue-related tasks.
Given the fact that different speaker roles behave differently, DSTC4 is a suitable benchmark dataset for evaluation.

We choose a mini-batch \texttt{adam} as the optimizer with the batch size of 128 examples~\cite{kingma2014adam}.
The size of each hidden recurrent layer is 128.
We use pre-trained 200-dimensional word embeddings $GloVe$~\cite{pennington2014glove}.
We only apply 30 training epochs without any early stop approach. 
The sentence encoder is implemented using a CNN with the filters of size $[2, 3, 4]$, 128 filters each size, and max pooling over time.
The idea is to capture the most important feature (the highest value) for
each feature map. This pooling scheme naturally
deals with variable sentence lengths. Please refer to \newcite{kim2014convolutional} for more details.

For both tasks, we focus on predicting multiple labels including speech acts and attributes, so the evaluation metric is average F1 score for balancing recall and precision in each utterance.
Note that the final prediction may contain multiple labels.

\subsection{Results}

The experiments are shown in Table~\ref{tab:res}, where we report the average number over five runs.
The first baseline (row (a)) is the best participant of DSTC4 in IWSDS 2016~\cite{kim2016fourth}, the poor performance is probably because tourist intents are much more difficult than guide intents (most systems achieved higher than 60\% of F1 for guide intents but lower than 50\% for tourist intents).
The second baseline (row (b)) models the current utterance without contexts, performing 62.6\% for understanding and 63.4\% for policy learning.

\subsubsection{Language Understanding Results}
\label{ssec:luresults}
With contextual history, using ground truth semantic labels for learning history summary vectors greatly improves the performance to 68.2\% (row (c)), while using natural language slightly improves the performance to 64.2\% (row (e)).
The reason may be that NL utterances contain more noises and the contextual vectors are more difficult to model for LU.
The proposed role-based contextual models applying on semantic labels and NL achieve 69.2\% (row (d)) and 65.1\% (row (f)) on F1 respectively, showing the significant improvement all model without role modeling.
Furthermore, adding the intermediate guidance acquires additional improvement (65.8\% from the row (g)).
It is shown that the semantic labels successfully guide the sentence encoder to obtain better sentence-level representations, and then the history summary vector carrying more accurate semantics gives better performance for understanding.

\subsubsection{Dialogue Policy Learning Results}
\label{ssec:policyresults}
To predict the guide's next actions, 
the baseline utilizes intent tags of the current utterance without contexts (row (b)).
Table~\ref{tab:res} shows the similar trend as LU results, where applying either role-based contextual models or intermediate guidance brings advantages for both semantics-encoded and NL-encoded history.

\subsection{Discussion}
In contrast to NL, semantic labels (intent-attribute pairs) can be seen as more explicit and concise information for modeling the history, which indeed gains more in our experiments for both LU and dialogue policy learning.
The results of Contextual-Sem can be treated as the upper bound performance, because they utilizes the ground truth semantics of contexts.
Among the experiments of Contextual-NL, which are more practical because the annotated semantics are not required during testing, the proposed approaches achieve 5.1\% and 6.3\% relative improvement compared to the baseline for LU and dialogue policy learning respectively.

Between LU and dialogue policy learning tasks, most LU results are worse than dialogue policy learning results.
The reason probably is that the guide has similar behavior patterns such as providing information and confirming questions etc., while the user can have more diverse interactions.
Therefore, understanding the user intents is slightly harder than predicting the guide policy in the DSTC4 dataset.

With the promising improvement for both LU and dialogue policy learning, the idea about modeling speaker role information can be further extended to various research topics in the future.

\section{Conclusion}
This paper proposes an end-to-end role-based contextual model that automatically learns speaker-specific contextual encoding.
Experiments on a benchmark multi-domain human-human dialogue dataset show that our role-based model achieves impressive improvement in language understanding and dialogue policy learning, demonstrating that different speaker roles behave differently and focus on different goals.

\section*{Acknowledgements}
We would like to thank reviewers for their insightful comments on the paper.
The authors are supported by the Ministry of Science and Technology of Taiwan and MediaTek Inc..

\bibliography{ijcnlp2017}
\bibliographystyle{ijcnlp2017}

\end{document}